\documentclass[twoside,leqno,twocolumn]{article}

\usepackage[letterpaper]{geometry}

\usepackage{ltexpprt}
\usepackage{hyperref}
\usepackage{graphicx}
\usepackage{longtable}
\usepackage{cite}
\usepackage{caption}
\usepackage{subfigure}
\usepackage{xcolor}
\usepackage{longtable}
\usepackage{supertabular,booktabs}
\usepackage{amsmath}
\usepackage{amssymb}
\usepackage{xspace}
\usepackage{comment}
\usepackage{multirow}
\usepackage{balance}
\usepackage{enumitem}

\begin{document}

\newcommand\relatedversion{}
\newcommand\attrib[1]{{\tt \small #1}\xspace}

\newcommand{\hadis}[1]{\textcolor{red}{(hadis) #1}}
\newcommand{\francesco}[1]{\textcolor{blue}{(francesco) #1}}
\newcommand{\abol}[1]{\textcolor{green}{(abol) #1}}
\newcommand{\nazanin}[1]{\textcolor{teal}{(Nazanin) #1}}
\newcommand{\new}[1]{\textcolor{black}{ #1}}

\title{FairPilot: An Explorative System for Hyperparameter Tuning \\ through the Lens of Fairness\thanks{This project was supported in part by the Institute of Education Sciences (R305D220055) and the National Science Foundation (NSF 2107290).}}


\author{
Francesco Di Carlo\thanks{University of Illinois Chicago. (fdicar2@uic.edu)}
\and Nazanin Nezami\thanks{University of Illinois Chicago.(nnezam2@uic.edu)}
\and Hadis Anahideh\thanks{University of Illinois Chicago. (hadis@uic.edu)}
\and Abolfazl Asudeh\thanks{University of Illinois Chicago. (asudeh@uic.edu)}}

\date{}

\maketitle


\fancyfoot[R]{\scriptsize{Copyright \textcopyright\ 2023 by SIAM\\
Unauthorized reproduction of this article is prohibited}}





\begin{abstract} 

Despite the potential benefits of machine learning (ML) in high-risk decision-making domains, the deployment of ML is not accessible to practitioners, and there is a risk of discrimination.
To establish trust and acceptance of ML in such domains, democratizing ML tools and fairness consideration are crucial. 
In this paper, we introduce FairPilot, an interactive system designed to promote the responsible development of ML models by exploring a combination of various models, different hyperparameters, and a wide range of fairness definitions. We emphasize the challenge of selecting the ``best" ML model and demonstrate how FairPilot allows users to select a set of evaluation criteria and then displays the Pareto frontier of models and hyperparameters as an interactive map. 
FairPilot is the first system to combine these features, offering a unique opportunity for users to responsibly choose their model.




\end{abstract}

\section{Introduction}\label{sec:intro}

Predictive analytics and machine learning (ML) have become increasingly prevalent in sensitive decision-making domains, such as healthcare, finance, criminal justice, and education. These domains involve high-stakes decisions with potentially significant impacts on people's lives \cite{giordano2021accessing,nieto2019usage,lo2020ethical}. For instance, in the education sector, predictive analytics and ML can be used to predict student outcomes, identify at-risk students, and provide personalized learning plans.
However, the use of these technologies in sensitive decision-making domains is a complex and multifaceted issue. 

While ML holds a significant promise for high-risk domains, its acceptability among practitioners can be influenced by several factors. Two key factors are the lack of democratization in deployment and the potential to exacerbate inequalities.
The former refers to the effort to make ML tools and technology widely accessible to a diverse group of users with different expertise and background. 
The latter refers to the fact that ML tools can perpetuate biases and discrimination if not designed and implemented carefully. For example, a model that assigns lower probabilities of success to students from certain demographic groups (such as students of color or students from low-income backgrounds) could exacerbate existing inequalities and result in unfair treatment for these students. This can create a disconnection between the use of ML and the values of equity and inclusion that are central to the mission of such sensitive domains (e.g., education).

Consider a domain practitioner who aims to develop a machine-learning model for a prediction task.
Recent advances in AI have introduced a wide range of choices for each ML model that can be employed to make predictions.
Having multiple options is beneficial as it gives the flexibility of choice to the data scientist. On the other hand, it is not clear which model is ``better'' for the given task. Therefore, exploring different options before making the final selection is cumbersome:
\begin{enumerate}
    \item {\em Hyperparameters}: ML models are often associated with various hyperparameters that require predefined values before training the model. The selection of hyperparameters may highly affect the developed model and its performance. The combination of different values for hyperparameters makes the exploration space for each model {\em exponentially large} to the number of its hyperparameters. 
    \item {\em Time complexity}: Training ML models is resource hungry task that is coupled with a large number of ML models to be trained and the exponential space of hyperparameters choices, makes the responsible model selection process overwhelming, particularly for practitioners.
    \item {\em Multiple evaluation criteria}:
    {Model selection solely based on maximizing the accuracy is not enough}, at least for sensitive decision-making environments since there often exists a trade-off between the fairness and (accuracy) performance of a model~\cite{bertsimas2012efficiency}. On top of that, fairness is not a unique notion; there are many (more than 21) fairness definitions~\cite{verma2018fairness,barocas2017fairness}, and those are often in trade-off with each other~\cite{kleinberg2016inherent}. Even more unwieldy, the relationships between the fairness notions (and model performance) are context-specific and are not clear apriori~\cite{anahideh2021choice}.
    Therefore, a data scientist is unlikely to be able to come up with a formula to combine all fairness and model performance criteria into a single objective.
\end{enumerate}

Responsible model development without the aid of {\bf assistive exploratory tools} is formidable, if not infeasible, task for data scientists. 
Conversely, despite extensive advances by the Fair ML community, there is a research gap and a desperate need for {\em interactive} explorative systems to help data-scientist practitioners develop ML models responsibly {~\cite{asudeh2021enabling,lewis2021teaching,arrieta2020explainable}.}

This paper is an attempt towards filling this gap. To the best of our knowledge, {\bf FairPilot} is the first \underline{interactive} system that enables users to explore a combination of (a) various \underline{ML models} and (b) different \underline{hyperparameter combinations} while considering (c) a \underline{wide range of fairness definitions}. FairPilot allows the user to select a set of evaluation criteria and shows the Pareto frontier of models and hyperparameters as an interactive map to the users allowing them to choose a proper model or explore other combinations of evaluation criteria before responsibly finalizing their choice. 

\new{The Pareto frontier represents the optimal trade-offs between various fairness measures and accuracy in a set of models. Users can visually compare models on the Pareto frontier and choose the model that best fits their needs and preferences, depending on the specific application. For example, if the decision-making process disproportionately affects certain racial or gender groups, the preference may be for the model with the highest fairness (based on a specific metric), even if it comes at the expense of some accuracy. Conversely, in other applications, accuracy may be the more important consideration, and the model that provides the highest accuracy may be preferred even if it has lower fairness. Ultimately, the choice of model will depend on the specific application and the relative importance of each objective.
}

Technically speaking, the core idea underlying FairPilot is that the space of the model/hyperparameter combinations is independent of the space of model evaluation criteria. This enables designing a grid-search algorithm over the full factorial of the model and hyperparameters during the {\em preprocessing time} and collecting (and indexing) comprehensive information that creates an interactive environment for the user exploration phase.

The paper is structured as follows. Section ~\ref{sec:related} provides a brief summary of the existing related studies. Section ~\ref{sec:prelim} provides technical background on Pareto Frontier Multi-objective optimization problems and how it is adapted for the FairPilot context. In sections, ~\ref{sec:system} and \ref{sec:ui}, the architecture and user interface options provided by FairPilot are explained. Section~\ref{sec:cc} presents a case study on the \emph{ELS} dataset to explore the hyperparameter space with FairPilot for model selection.

\section{Related Work}\label{sec:related}

The democratization and accessibility of machine learning have been a rapidly growing area of research in recent years. Many scholars have investigated methods and tools to make machine learning more accessible to non-experts and to promote more equitable access to the benefits of this technology. An open-source machine learning platform that allows non-experts to build and deploy predictive models has been proposed in \cite{van2021open}. Many others have proposed visualization techniques to help users better understand and interpret machine learning results \cite{chatzimparmpas2020survey}. 
Another area of research has been the development of automated machine learning (AutoML) tools, which can automate many of the tasks typically performed by machine learning experts. AutoML tools, such as those presented in \cite{feurer2015efficient}, can make it easier for non-experts to create effective ML models without needing to understand the complexities of the underlying algorithms and techniques.

{Most ML models feature a set of hyperparameters that must be predefined for training, and the choice of these hyperparameters significantly affects the model's ability to make correct predictions \cite{claesen2015hyperparameter,weerts2020importance}. Therefore, several studies focus on constructing grid or random search strategies \cite{bergstra2012random,yu2020hyper}, or optimizing for the set of hyperparameters using sequential learning techniques (e.g., Bayesian optimization) \cite{yu2020hyper,turner2021bayesian}. However, an accurate model can be biased and unfair in prediction since inherent trade-offs exist in using ML for decisions that impact social welfare.}

{Fairness in ML involves a growing body of research dedicated to identifying and addressing biases in algorithms \cite{mehrabi2021survey, barocas2017fairness}. Fair-ML promotes the idea that algorithms should not produce biased or discriminatory outcomes and that their decisions should be impartial and equitable for all individuals, groups, or intersectional subgroups \cite{john2020verifying,foulds2020intersectional}. On the other hand, a fair model can be inaccurate and useless in practice. Therefore, several studies aimed at designing interventions to balance the trade-off between fairness and accuracy in predictive modeling tasks \cite{kamiran2012data,vzliobaite2015relation,liu2022accuracy}.}

One resolution to handle the fairness and accuracy trade-off is attainable by designing a fairness-aware hyperparameter search approach. A fair hyperparameter-tuning tool promotes consideration of fairness in the choice of hyperparameter for model selection by measuring the fairness metrics once the model has been trained and optimized for accuracy. \cite{rolf2020balancing} considers algorithmic policies such as hyperparameters for balancing the social welfare and private objective (e.g., profit) based on individual fairness. \cite{perrone2021fair} designs a generally constrained Bayesian optimization framework and reveals that accurate and fair solutions are achievable by acting solely on the hyperparameter. \cite{cruz2021promoting} provides a simple and flexible intervention to incorporate fairness objectives in ML pipelines by proposing the fair variants of 
hyperparameter optimization algorithms such as Fair Random Search, Fair TPE, and Fairband.




\section{Preliminaries}\label{sec:prelim}
{Fairness-aware exploration on the space of hyperparameters of machine learning models can be considered as a Multi-objective optimization problem where accuracy and fairness are conflicting objectives.

Multi-objective optimization \cite{marler2004survey, ngatchou2005pareto} considers problems with more than one objective function to be optimized simultaneously.
Two popular directions for multi-objective optimization are 1) combining the optimization criteria in form of a single function (Equation~\ref{eq:objfunc}) and 2) identifying the Pareto frontier.

\begin{align}\label{eq:objfunc}&
\max_{\mathbf{x} \in \mathcal{X}} \mathbf{f}(\mathbf{x})
\end{align}

In Equation~\ref{eq:objfunc} $\mathcal{X} \in \mathbb{R}^d$ is the bounded search space such that $\mathbf{x}^L \leq \mathbf{x} \leq \mathbf{x}^U, \forall \mathbf{x} \in \mathcal{X}$ with $\mathbf{x}^L$ and $\mathbf{x}^U$ be the coordinate-wise lower and upper bounds, and  $\mathbf{f}: \mathcal{X} \rightarrow \mathbb{R}^M$ is the vector function of the $M$ multiple objectives, $\mathbf{f}=(f_1(\mathbf{x}),\dots, f_M(\mathbf{x}))$.

In the context of FairPilot, the hyperparameter tuning problem can be considered as a black-box function $\mathbf{f}(\mathbf{x})$, where the accuracy $f_A(\mathbf{x})$ and fairness $f_F(\mathbf{x})$ objectives are defined over a set of input hyperparameters $\mathbf{x} \in \mathcal{X}$. This implies that the task of selecting the optimal model requires considering multiple objectives, each with its own set of hyperparameters, making it a complex optimization problem.
Due to the conflicting nature of objectives,
there is no unique solution that optimizes all combination functions $\mathbf{f}$. 
In such environments specifying a meaningful function $\mathbf{f}$ is challenging and often not practical for ordinary users.

Therefore, our aim is to find a set of equally desirable solutions that hold a trade-off between the objectives, known as \emph{Pareto Frontier}. The Pareto set contains only dominant (aka non-dominated) solutions that cannot be strictly dominated by any other solutions 
in at least one objective \cite{furlan1908vilfredo}.

Suppose we aim to explore the space of hyperparameters for solving a binary classification task where the target variable $Y \in \{0,1\}$ defines 1 for positive, and 0 for negative outcomes, and $S \in \{0,1\}$ is the sensitive attribute (e.g., race) represent two different population subgroups (e.g., White and Black). Now, for a given set of input hyperparameters, the accuracy metric can be simply defined as the fraction of correct classification predictions or $f_A(\mathbf{x})=P(\hat{Y}=1 \vert Y=1)+P(\hat{Y}=0 \vert Y=0)$. Moreover, there are several fairness metrics that can be used to measure bias in predictive modeling for binary classification tasks such as \emph{statistical parity}, \emph{equal opportunity}, and \emph{predictive parity}.
For instance, \emph{statistical parity} metric is of the form $f_F(\mathbf{x})= P( \hat{Y}=1 \vert S = 0) - P(\hat{Y}= 1 \vert S = 1)$. 
Detailed explanations and formulas of various fairness metrics are provided in several fair ML studies \cite{verma2018fairness,mehrabi2021survey}.

FairPilot is an interactive exploration tool designed to find non-dominated hyperparameters given a dataset and user-defined model and fairness metric choices.
FairPilot outputs the set of non-dominated hyperparameters, $\mathcal{P}=\{\mathbf{x}^{(i)} \in \mathcal{X} \vert \nexists \mathbf{x^{'}} \in \mathcal{X}, f_F(\mathbf{x^{'}}) \leq f_F(\mathbf{x}^{(i)}),f_A(\mathbf{x^{'}}) \geq f_A(\mathbf{x}^{(i)}) \}$ that can be employed to train models in a fairness-aware manner. For each ML model, $\mathcal{P}$ reveals the set of hyperparameter values that can be used to train the model such that is strictly better than other models based on at least one of the unfairness $f_F$ or accuracy $f_A$ metrics. 


\section{System Specification}\label{sec:system}

\subsection{Architecture and Implementation}\label{sec:arch}

FairPilot is a web-based interactive system that explores the influence of hyperparameters on the trade-off between accuracy and fairness in predictive modeling. The goal is to support users in selecting the optimal combination of hyperparameters for training a model considering the fairness of the prediction outcome. FairPilot automatically trains models on a discretized space of user-defined hyperparameters and constructs the Pareto frontier by considering both accuracy and fairness metrics. 
The Pareto frontier contains the non-dominated solutions where the set of hyperparameters has produced the best results either from an accuracy or fairness perspective.
FairPilot is developed using \textit{Python 3.10.9}, and the web interface is created using \textit{Flask} framework. The system generates interactive plots using the \textit{Plotly} library.

\section{User Interface}\label{sec:ui}
The following section describes the main components of the FairPilot user interface (UI).

\begin{figure*}[ht!]
  \centering
  \includegraphics[width=0.8\linewidth]{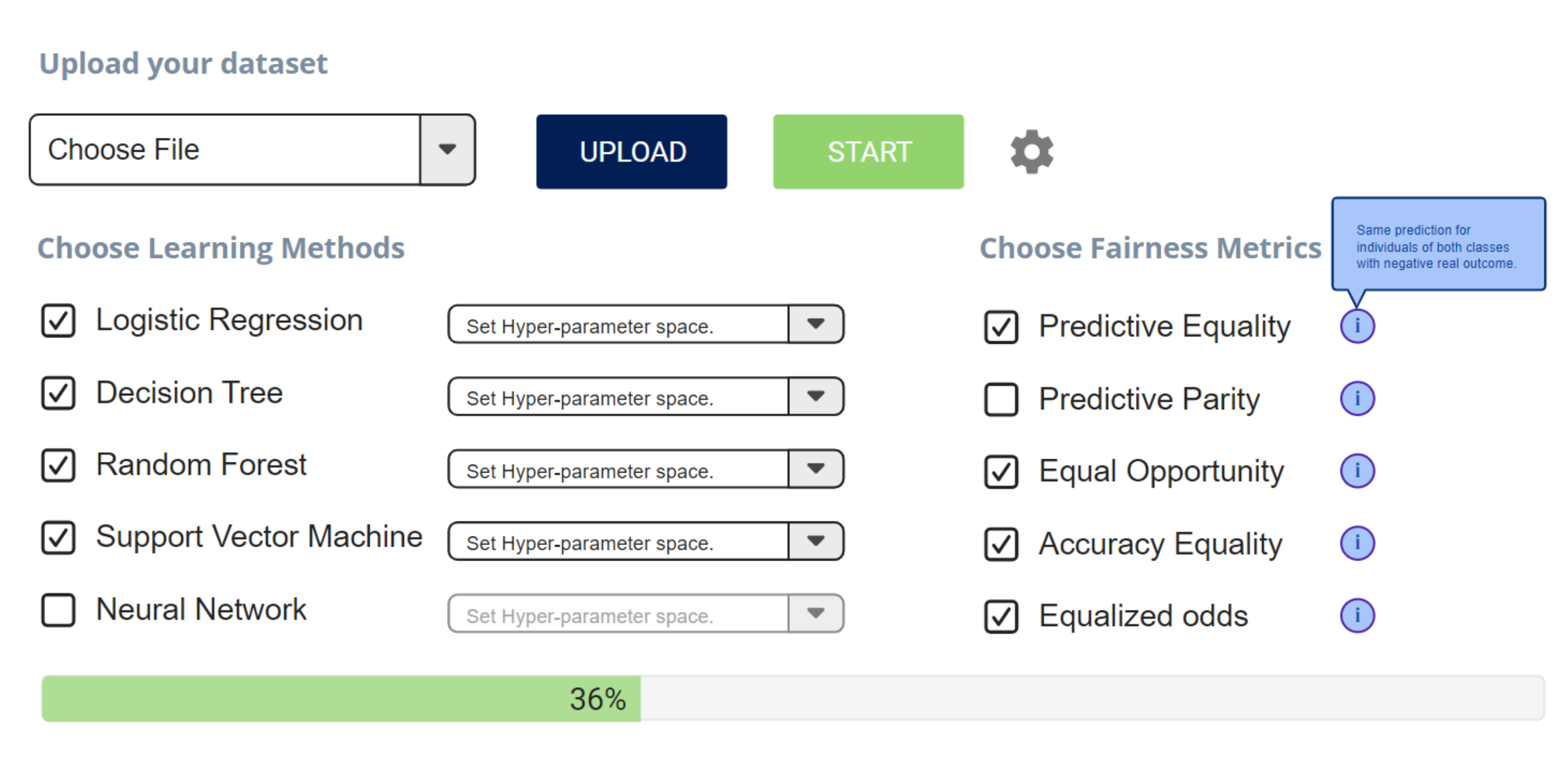}
  \caption{FairPilot User Interface.}
  \label{fig:user_interface}
\end{figure*}

\begin{figure*}[ht!]
\centering
       \subfigure[Selection of the Hyperparameter Space.]{\label{fig:2a}{\includegraphics[width=0.46\linewidth]{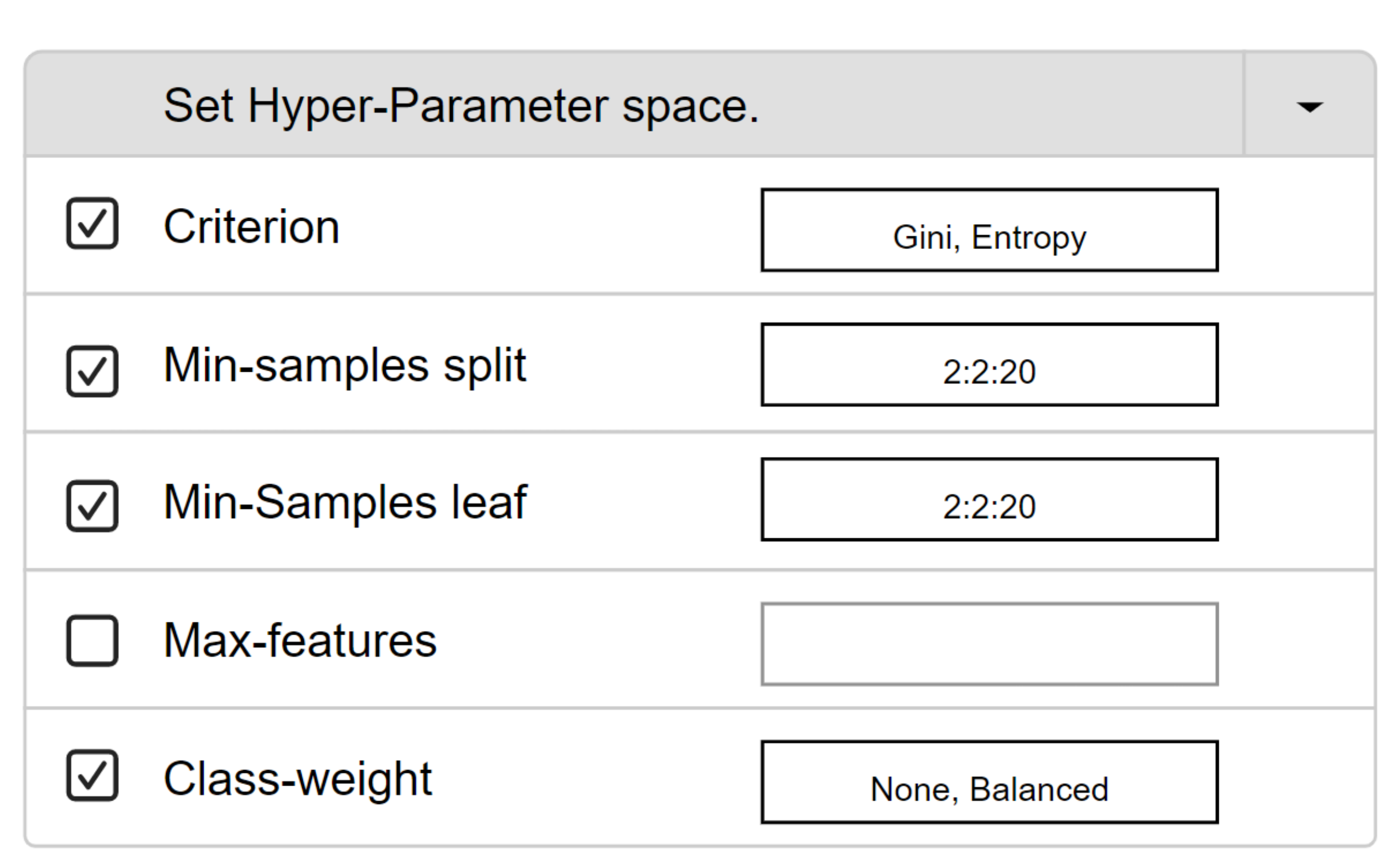}}} 
       \subfigure[Advanced settings.]{\label{fig:2b}{\includegraphics[width=0.40\linewidth]{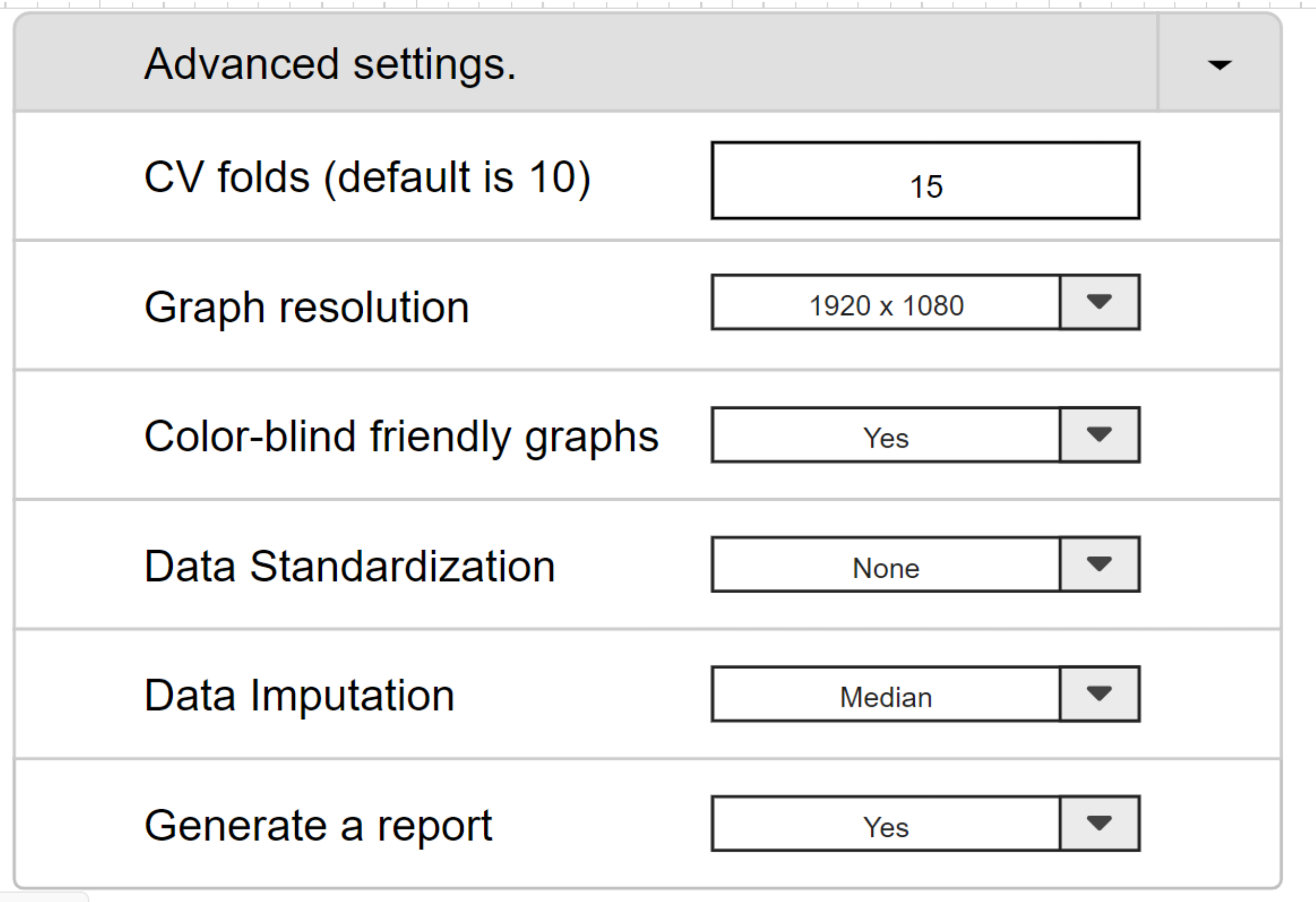}}}
  \caption{Drop down menus in the input sections.}
  \label{fig:HypSpace.pdf}
\end{figure*}


\subsection{Dataset Selection Section.}\label{sec:data}
The UI for FairPilot is shown in Figure \ref{fig:user_interface}. 
To begin using FairPilot, the user must first specify the dataset of interest. The dataset consists of a set of predictor variables and a categorical response variable. Once the dataset is uploaded, the user must select the target variable. The user is then asked to specify the sensitive attributes. 

\noindent\textbf{Learning Methods Section.}
Users can select from various learning models. The tool can handle a variety of machine learning models, ranging from basic models to more complex ones like Neural Networks.

\noindent\textbf{Hyper-Parameter Space Section.}
This section is displayed as a dropdown menu, as shown in Figure \ref{fig:2a}, where the user can manually input the hyperparameter space or use the default values. It is worth noting that the set of hyperparameters varies for each model, and our system provides default values for each. The backend of our system performs feasibility checks on user-defined values, and in case of improper range definition, a notification is sent to the user.

\noindent\textbf{Fairness Metrics Section.}
The user proceeds to the Fairness Metrics Section, where they can choose the fairness metrics of interest for their application. By clicking on the ``info'' button, a brief description of each metric is provided.
In our system, fairness is defined with regard to sensitive attributes, which describe social characteristics of individuals that are considered private, and can potentially result in discrimination when used in decision-making. These attributes are often related to aspects of an individual's identity, such as race, gender, and age.
Our tool can handle a broad range of fairness metrics, however, based on our previous research, certain group fairness metrics are highly positively or negatively correlated, while others are completely orthogonal \cite{anahideh2021choice}. 

\noindent\textbf{Advanced Settings Section.} 
Before performing the exploration, the user can define additional options by clicking on the gear wheel icon, as shown in Figure \ref{fig:2b}. These settings include:
\begin{itemize}[leftmargin=*]
    \item The number of cross-validation folds to use for each combination of hyperparameters.
    \item The resolution of the plots in case of export.
    \item The option to generate color-blind-friendly plots.
    \item The ability to perform various data standardization and imputation pre-processing techniques.
    \item An automated report generation for the results.
\end{itemize}


{FairPilot's default setting considers 5-fold cross-validation, average resolution, color-friendly plots, standard scalar, and removing NA values for data pre-processing with 
automated report generation enabled.}

\noindent\textbf{Loading Bar Section.} 
{After the user finishes setting up the desired options for the learning process, they can run the FairPilot tool by clicking on the Start button}. At this point, a loading bar will appear at the bottom of the page, as shown in Figure \ref{fig:user_interface}, to provide an estimate of the exploration progress. We utilize the \textit{tqdm} Python library to display the loading bar.

\subsection{Output Sections.}\label{sec:data} 
Once the learning process is complete, FairPilot generates the outputs as listed below. It is worth noting that the plots presented in this paper are resulted from analyzing the ELS dataset, which we shall elaborate on later in \S \ref{sec:cc}.

\begin{itemize}
     \item \textbf{Individual Model Pareto Frontiers.} FairPilot generates interactive plots that display an estimated Pareto frontier for each combination of learning methods and fairness metrics. An example is provided in Figure \ref{fig:outputs.pdf}. By hovering over any point on the plot, the user can see the exact hyper-parameter combination that produced that performance. Furthermore, colors, symbols, and sizes, as shown in Figure \ref{hover}, can be used to assist with interpretations (e.g., which hyper-parameters have the most significant impact on the final performance), and an interactive legend allows the user to focus on specific hyperparameters. 

     \item \textbf{Multi-Model Pareto Fronts.} FairPilot generates an interactive plot for each fairness metric that displays an estimated Pareto frontier for all of the ML methods combined, allowing the user to check the overall performance of the framework. Similar to the Individual Model Pareto Frontiers, the user can hover over any point on the plot to see the hyper-parameter combination that produced that performance. Different colors are used to identify different ML methods, and an interactive legend allows the user to focus on specific methods. An example is provided in Figure \ref{fig:3b}.
     
     \item \textbf{Superimposed Pareto Frontiers} FairPilot produces an interactive graph that superimposes the Pareto frontiers for each ML method, given a fairness metric. This graph allows the user to understand which learning method is the most promising overall and if any one method dominates the others. Figure \ref{fig:3c} provides a preview of this feature. A similar plot can be generated for an individual machine learning (ML) model using all fairness metrics. This interactive plot will display the estimated Pareto frontier for the ML model and allow the user to explore the impact of different fairness metrics on the model's performance. 

     \item \textbf{Superimposed Individual Model Pareto Frontiers} Given a single ML method, FairPilot produces an interactive graph that superimposes the Pareto frontiers for multiple fairness metrics. This can be used when the user is interested in more than one metric at a time. A preview is given in \ref{fig:smpf}.
     
    \item \textbf{Pareto Data Frames.} FairPilot generates data frames for each fairness metric that include all the Pareto points and their corresponding hyperparameter combinations. Additionally, the data frames allow the user to check the value of other fairness metrics (for example, on the data frame showing the Pareto points for the Predictive Equality metric, the user can also check the corresponding Equalized Odds values). Table \ref{tab:ppdt} provides a preview of this feature.
\end{itemize}

\new{Our system offers several interactive features that allow users to customize the visualizations and explore the data in more detail.
In addition to the pop-ups of information when hovering over data points, users can select groups of points on the legend to remove them from the visualization. They can also modify the hyperparameters that determine the size, color, and shape of the points in real time, allowing them to gain more insight into how changes to the hyperparameters affect model performance.
Furthermore, our system offers basic functionalities such as panning, zooming, fullscreen view, and the ability to download, which allow users to interact with the visualizations and explore the data in different ways.}



\section{Case Study}\label{sec:cc}

\textbf{Dataset.}
In this section, we present a case study to demonstrate how FairPilot can be used in practice. We utilize the Education Longitudinal Study of 2002 (ELS) dataset \footnote{ For more information about ELS, visit the main website: 
\url{https://nces.ed.gov/surveys/els2002/avail_data.asp}}), which is a longitudinal study designed to provide trend data about critical transitions experienced by students.
For our analysis, we consider the \emph{highest level of degree} as the target variable and create a binary classification task to predict whether an individual's highest degree earned is above or below a bachelor's degree.
To be specific, we assign a label of 1 to students who have obtained a college degree (i.e., a bachelor's degree or higher) as the favorable outcome, while students who have not achieved a college degree are assigned a label of 0 as the unfavorable outcome.
We consider the sensitive attribute of ``race'', which includes five groups of students: White, Black, Hispanic, Asian, and Multi-racial (MR). To evaluate fairness, we divide this attribute into two sub-categories: category 0 comprising Black, Hispanic, and MR students, and category 1 comprising Asian and White students.
This separation is based on precedent analysis of the ELS database in \cite{unknown}.
Data cleaning is performed to identify and rename the missing values (based on the documentation) and remove the observations that have many missing attributes ($>75\%$ of the attributes are missing).
The non-categorical data is then standardized by subtracting the mean and scaling to unit variance, but no imputation is performed. 
At this point, pre-processing is considered concluded and it is possible to proceed to model testing and training.

\noindent\textbf{Experimental setup.}
We evaluated four ML methods on the dataset: Decision Tree Classifier (DT), Random Forest Classifier (RF), Logistic Regression (LR), and Support Vector Classifier (SVC). Due to the need for interoperability in the education domain, we did not use Neural Network (NN), even though it is available in our tool.
The hyperparameters used for each ML method are presented below along with a brief description.




\textbf{Decision Tree Classifier:} The \attrib{criterion} is a function used to measure the quality of a split relative to the ideal case of perfect separation. The \attrib{min sample split} is the minimum number of samples required to split an internal node. The \attrib{minimum sample leaf} is the minimum number of samples required for a node to become a leaf. The \attrib{max features} is the number of features to consider when looking for the best split. The \attrib{class weight} is used to assign different weights to the classes in the dataset during the training process.

\textbf{Random Forest Classifier:} Same hyperparameters of the Decision Tree Classifier are used, with the addition of \attrib{bootstrap}, which is a boolean attribute that determines whether bootstrap samples are used when building trees. If False, the whole dataset is used to build each tree.

\textbf{Logistic Regression:} The hyperparameters used in logistic regression are \attrib{penalty}, which specify the norm of the penalty, and
\attrib{C}, which is the inverse of regularization strength.


\textbf{Support Vector Classifier:} Two main hyperparameters in SVC are \attrib{C}, which is the inverse of the regularization strength, and \attrib{kernel}, which specifies the kernel type to use in the SVC algorithm. 

Table \ref{tab:hs} shows the range of the attributes that define the hyperparameter space for different considered ML models.
\begin{table}[htbp]
  \centering
  \scriptsize
  \caption{Hyperparameters Space}
    \begin{tabular}{|c|ll|}
    \toprule
    \multicolumn{1}{|l|}{Model} & Hypeparameter & Range \\
    \midrule
    \multicolumn{1}{|c|}{\multirow{5}[2]{*}{Decision Tree}} & criterion & [gini, entropy] \\
          & max features & [None, sqrt, log2] \\
          & min samples split & [2, 4, 8, 12, 16, 20] \\
          & min samples leaf & [1, 4, 8, 12, 16, 20] \\
          & class weight & [None, balanced] \\
    \midrule
    \multirow{6}[2]{*}{Random Forest} & criterion & [gini, entropy] \\
          & max features & [None, sqrt, log2] \\
          & min samples split & [2, 4, 8, 12, 16, 20] \\
          & min samples leaf & [1, 4, 8, 12, 16, 20] \\
          & class weight & [None, balanced] \\
          & bootstrap & [False, True] \\
    \midrule
    \multirow{3}[2]{*}{Logistic Reg.} & C     & [0.001, 0.01, 0.1, 1,  \\
          &       & 10, 100, 1000] \\
          & penalty & [l2, none] \\
    \midrule
    \multirow{3}[2]{*}{SVC} & C     & [0.001, 0.01, 0.1, 1, \\
          &       &  10, 100, 1000] \\
          & kernel & [linear, poly, rbf, sigmoid] \\
    \bottomrule
    \end{tabular}%
  \label{tab:hs}%
\end{table}%


\noindent\textbf{Evaluation.} 
In addition to \emph{accuracy}, we evaluate the performance of our models using five fairness metrics, namely: \emph{predictive parity}, \emph{predictive equality}, \emph{equal opportunity}, \emph{accuracy equality}, and \emph{equalized odds}. The evaluation is conducted using 10 different data splits for training and testing. We collect the mean and variance of both accuracy and fairness metrics across all 10 runs. For each ML model and hyperparameter combination, we train and test the models through a brute-force approach. We then construct the Pareto Front using the trained models using the two dimensions of fairness and accuracy. We use the Pareto Front to identify the optimal model settings.


\noindent\textbf{Results.}
FairPilot is an assistive tool for exploring the modeling space and investigating the trade-off between accuracy and fairness. It serves two distinct objectives: model interpretation and model selection.
For model interpretation, in Figure \ref{fig:3a}, which displays the \emph{accuracy} vs {equal opportunity} Pareto front when using a Random Forest. We can observe that the \attrib{bootstrap} is an influential hyperparameter in determining the performance of the model. When \attrib{bootstrap}=True, the fitted models generally have higher accuracy and lower \emph{equal opportunity}. FairPilot provides a customizable color-coding and allows the user to select the hyperparameter of interest, resulting in reports that align with their perspective and facilitate a better interpretation of the results.
In terms of model selection, the user can check the best possible configurations at a glance,  by hovering on the Pareto front points as shown in Figure \ref{hover}, for which we use the Decision Tree classifier and \emph{predictive equality} for fairness evaluation.  
Alternatively, the user can examine various configurations on the corresponding Pareto Data Frame (as shown in Table \ref{tab:ppdt}). This makes it effortless and intuitive for the user to select the appropriate ML model for their specific application.

\begin{figure*}[!ht]
\centering
\subfigure[Individual Pareto front, Accuracy vs Predictive Equality using Random Forest.]{\label{fig:3a}{\includegraphics[width=0.45\linewidth]{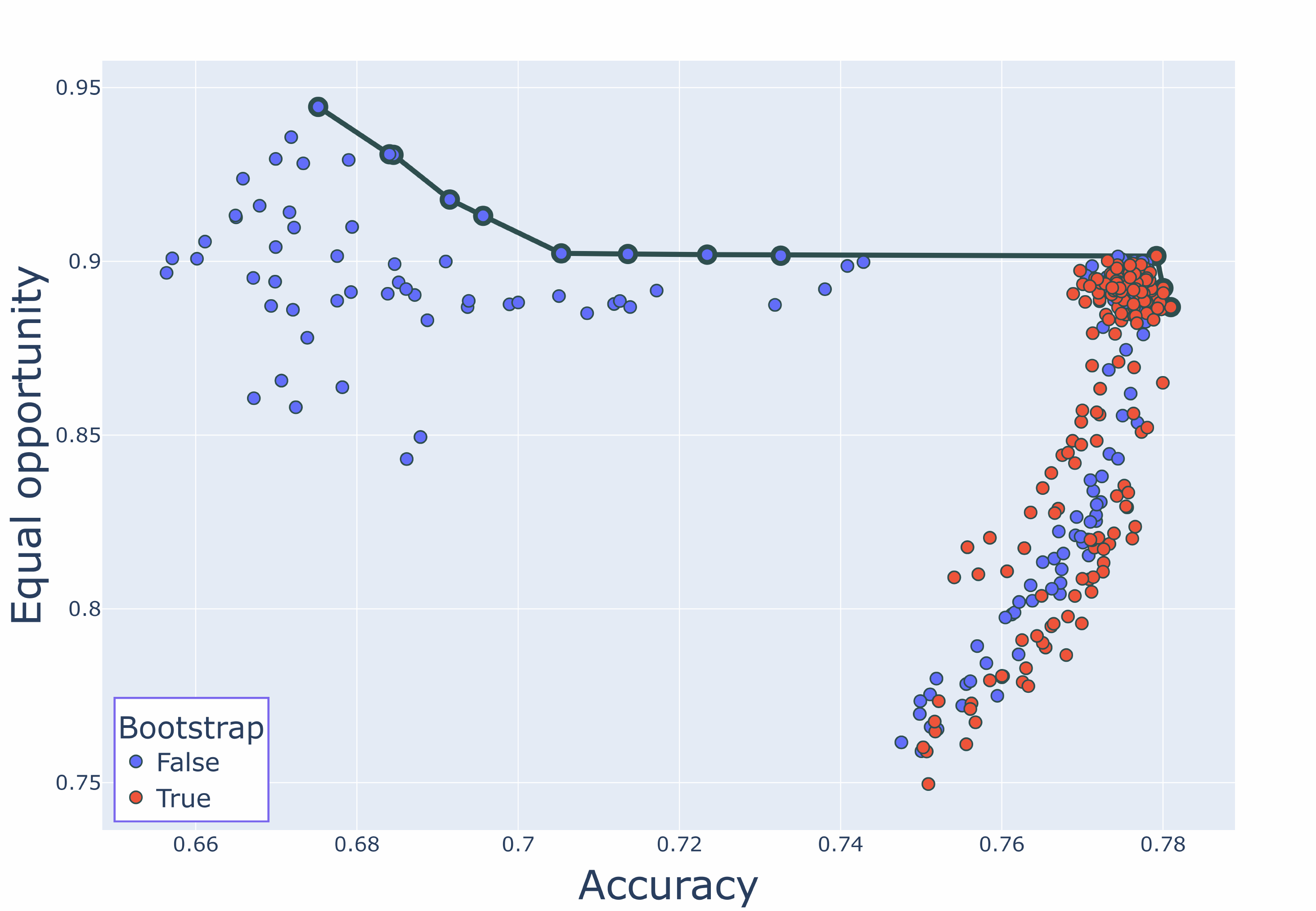}}}
\hspace{1 mm}
\subfigure[Multi-Model Pareto front, Accuracy vs Predictive Equality.]{\label{fig:3b}\includegraphics[width=0.45\linewidth]{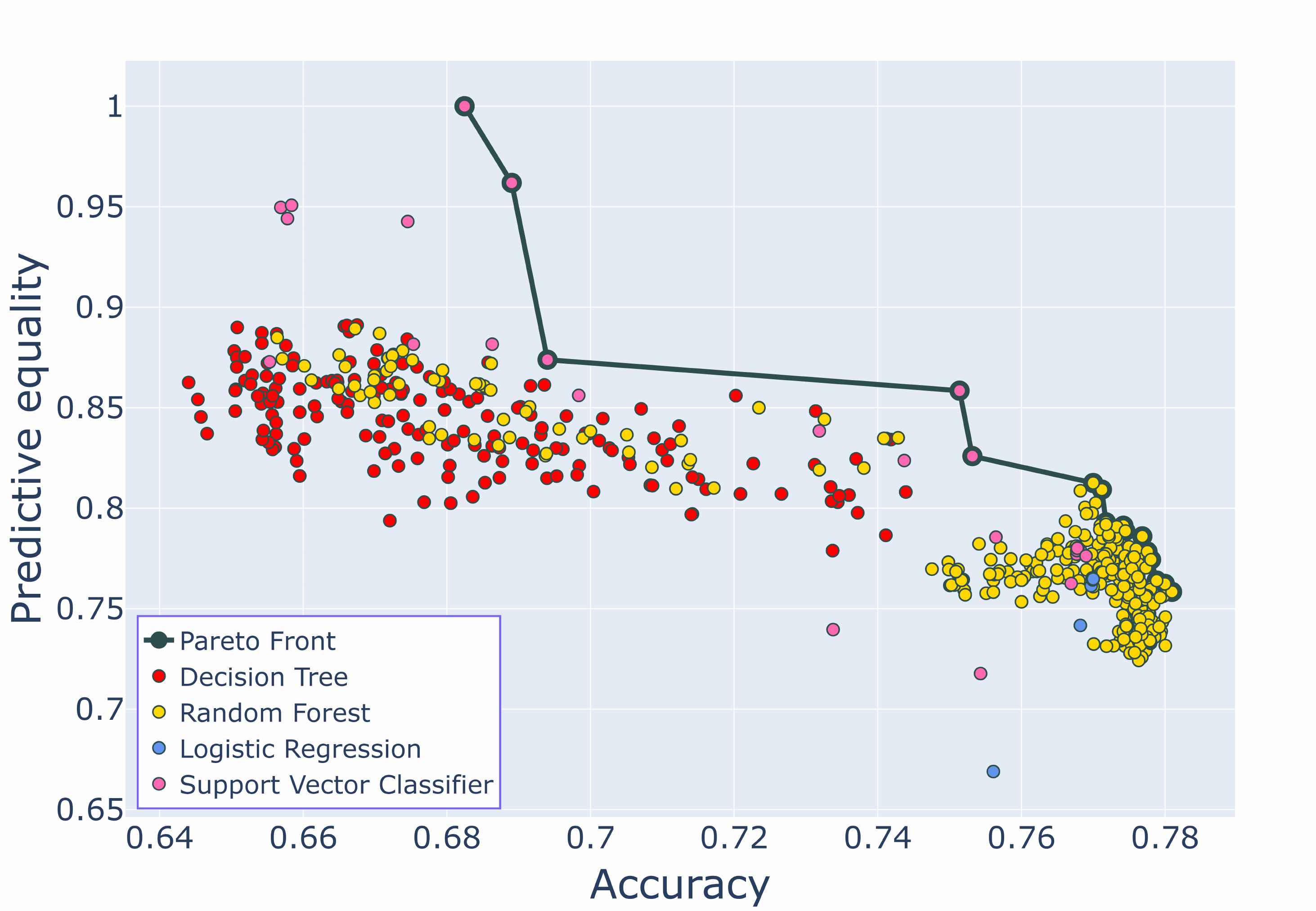}}

\subfigure[Superimposed Pareto Fronts, Accuracy vs Equalized Odds.]{\label{fig:3c}\includegraphics[width=0.45\linewidth]{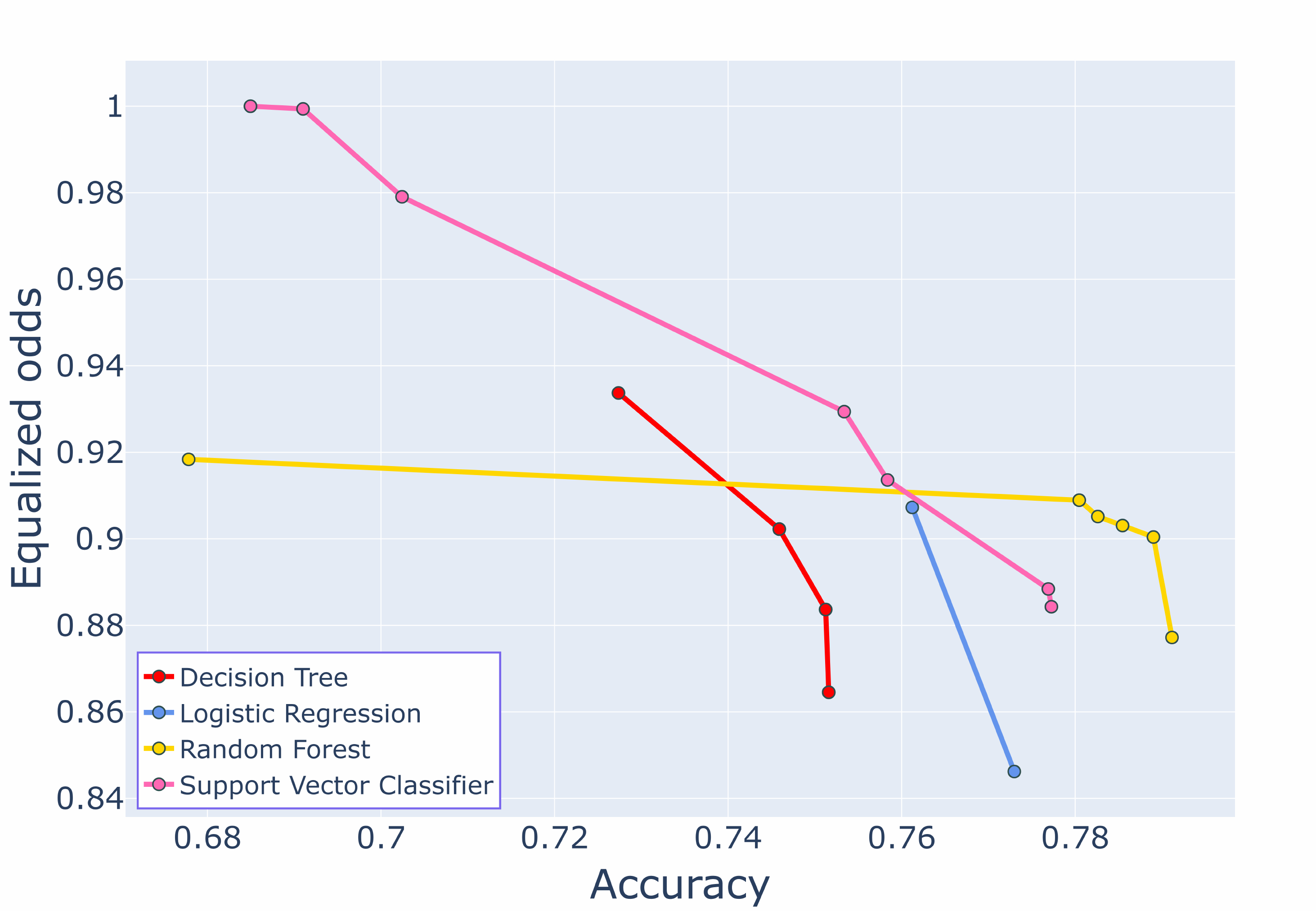}}
\hspace{1 mm}
\subfigure[Single model Pareto fronts, Accuracy vs Multiple metrics using a Decision Tree.]{\label{fig:smpf}\includegraphics[width=0.45\linewidth]{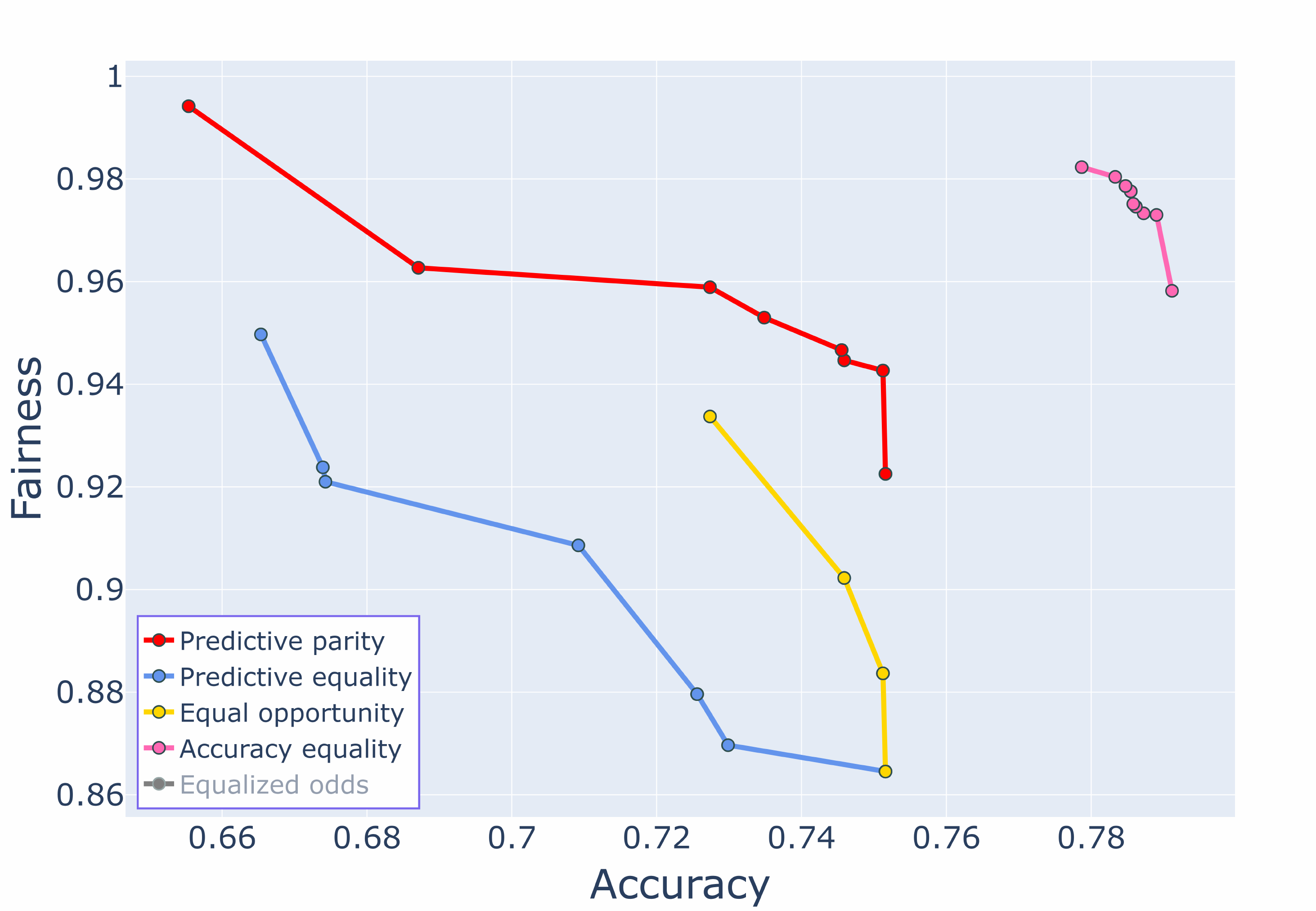}}

\subfigure[Individual Pareto Front, Demo of the interactive graph using Predictive Parity and a Decision Tree.]{\label{hover}\includegraphics[width=0.45\linewidth]{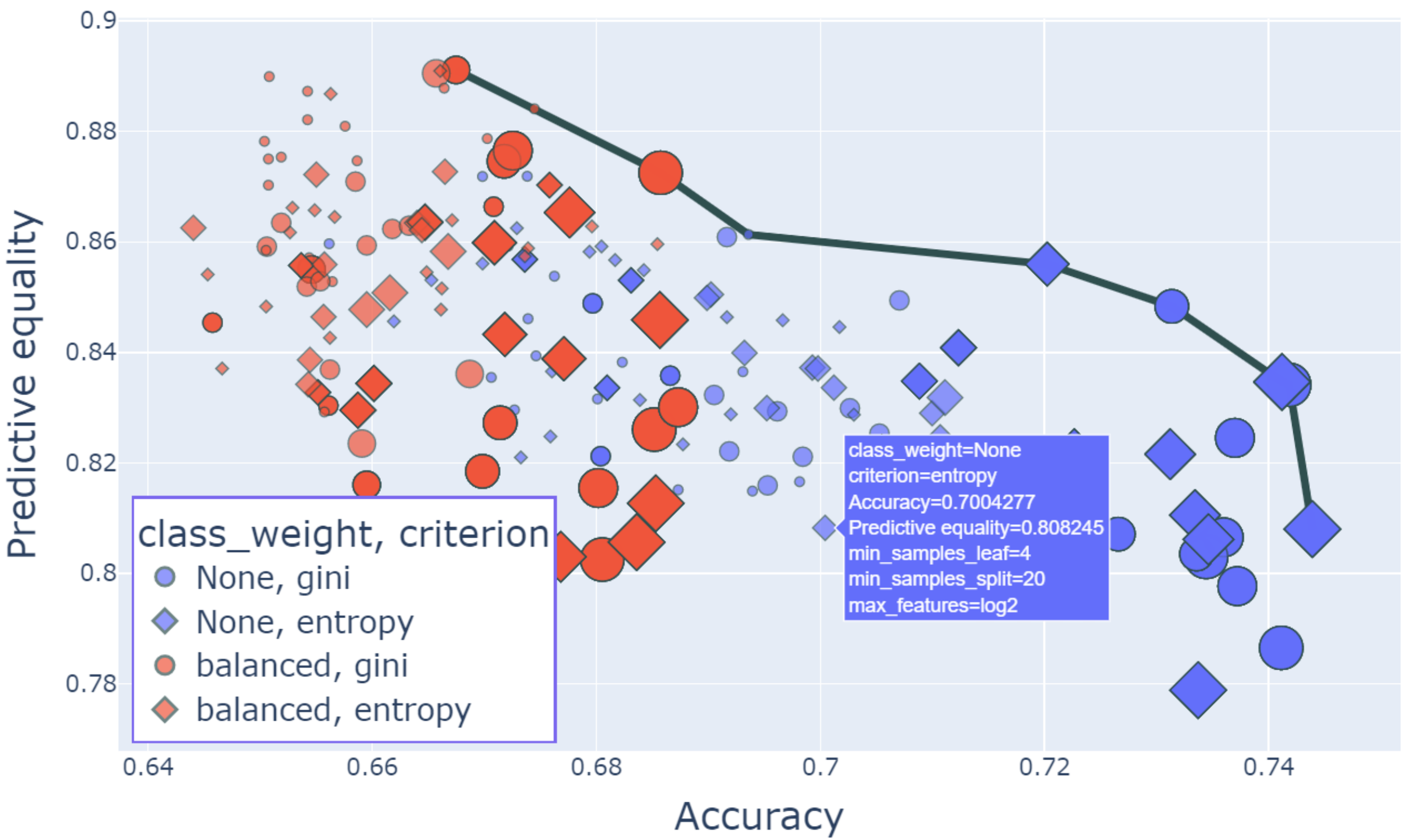}}

  
\caption{Output plots of FairPilot}
\label{fig:outputs.pdf}
\vspace{-4mm}
\end{figure*}




\begin{table*}[t]
  \centering
  \scriptsize
  \caption{Pareto points for Accuracy vs Predictive Parity, using a Decision Tree}
    \begin{tabular}{|c|c|c|c|c|c|c|c|c|c|}
        \hline
    \toprule
    \multicolumn{1}{|c|}{\textbf{Accuracy}} & \multicolumn{1}{|c|}{\textbf{Pred. Par.}} & \multicolumn{1}{|c|}{\textbf{Pred. Eql.}} & \multicolumn{1}{|c|}{\textbf{Eql. Opp.}} & \multicolumn{1}{|c|}{\textbf{Acc. Eql.}} & \multicolumn{1}{|c|}{\textbf{min\_leaf}} & \multicolumn{1}{|c|}{\textbf{min\_split}} & \multicolumn{1}{|c|}{\textbf{max\_features}} & \multicolumn{1}{|c|}{\textbf{Obj.}} & \multicolumn{1}{|c|}{\textbf{clf. weights}} \\
    \hline
    0.655 & 0.741 & 0.741 & 0.754 & 0.950 & 16    & 2     & log2  & entropy & balanced \\
    0.687 & 0.801 & 0.801 & 0.799 & 0.944 & 20    & 2     & None  & entropy & balanced \\
    0.727 & 0.788 & 0.788 & 0.934 & 0.987 & 8     & 20    & log2  & gini  & None \\
    0.735 & 0.777 & 0.777 & 0.868 & 0.955 & 12    & 2     & log2  & entropy & None \\
    0.746 & 0.819 & 0.819 & 0.876 & 0.950 & 20    & 2     & None  & entropy & None \\
    0.746 & 0.831 & 0.831 & 0.902 & 0.959 & 20    & 2     & sqrt  & entropy & None \\
    0.751 & 0.828 & 0.828 & 0.884 & 0.949 & 20    & 2     & None  & gini  & None \\
    0.752 & 0.865 & 0.865 & 0.865 & 0.924 & 12    & 2     & None  & gini  & None \\\hline
    \end{tabular}%
  \label{tab:ppdt}%
  \vspace{-4mm}
\end{table*}%

Figure \ref{fig:3b} shows \emph{accuracy} vs \emph{predictive equality} Pareto front considering all different ML models. The results indicate that SVC achieves the best performance in terms of the \emph{predictive equality}, while the Random Forest classifier tends to be more accurate. Once again, a clear trade-off is observed. Comparing the model with the highest accuracy against the model with the highest fairness, we have to sacrifice $10\%$ in accuracy to achieve a $25\%$ increase in fairness.
The user has the option to choose any other point on the Pareto front, based on the application and the desired level of emphasis on accuracy and fairness.

In Figure \ref{fig:3c}, which displays the superimposed Pareto fronts of each ML method taken individually when considering \emph{accuracy} vs \emph{equalized odds}, we can observe that no single ML method dominates over the others. SVC is capable of producing fairer results, while RF has better accuracy. However, a trade-off exists in each Pareto Front, indicating that there is no single best model for all situations. 

{Figure \ref{fig:smpf} shows the Pareto front of accuracy and fairness for the DT model using various fairness metrics. The results indicate that sacrificing accuracy for fairness is more achievable for the \emph{equal opportunity} metric (yellow line) compared to other metrics. Notably, by comparing the endpoints of \emph{equal opportunity} Pareto front, we observe that a $7\%$ increase in fairness can be attained with a mere $3\%$ loss in accuracy.
However, the importance assigned to fairness and accuracy in the model selection process may differ for each user when considering other fairness metrics.}

Figure \ref{hover} demonstrates the level of personalization, interactivity, and density of information provided by the plots in FairPilot. 
This graph allows the user to immediately observe the effects of three hyperparameters at the same time: \attrib{class weight} is represented using a color code, \attrib{criterion} is indicated using different symbols, and \attrib{min samples leaf} is shown by varying the size and transparency of the markers. By hovering over any point on the graph, the user access the exact configuration used, along with the corresponding objective values. As we can see, a 'balanced' class weight is able to obtain generally fairer models while accuracy is higher if we set the class weight as equal for both classes.

\section{Conclusion and Future Work}

This paper has presented FairPilot, an innovative solution for addressing the challenges associated with deploying machine learning models in high-risk decision-making domains while promoting fairness. FairPilot allows users to explore various models, hyperparameters, and fairness definitions and displays the Pareto frontier of models and hyperparameters as an interactive map. This unique combination of features offers users an opportunity to responsibly choose their ML models based on their application and objectives.
FairPilot's ability to explore and visualize the Pareto frontier of models and hyperparameters enables users to make informed decisions and trade-offs between accuracy and fairness. The tool also enables users to explore the impact of various hyperparameters on model performance and fairness, which can be especially useful when dealing with complex models.

\new{In the future, we plan to expand FairPilot's range of fairness definitions and integrate new models and hyperparameters. We aim to incorporate state-of-the-art algorithms such as Multi-Objective Bayesian Optimization (MOBO) to enable efficient hyperparameter optimization and to make FairPilot compatible with larger datasets.
We also plan to enhance the tool's ability to deal with non-binary sensitive features and multiple sensitive features simultaneously. We believe that this will enable FairPilot to be more widely applicable in a range of decision-making domains.}

\new{Furthermore, we aim to improve the accessibility and ease of use of FairPilot for practitioners. This includes incorporating user-friendly interfaces and workflows that allow practitioners to interact with the tool easily and providing documentation and tutorials to guide them through the use of FairPilot.
Overall, we believe that FairPilot has the potential to significantly impact responsible AI and decision-making practices by enabling informed decisions based on both model performance and fairness criteria.}



\newpage
\bibliographystyle{siam}
\bibliography{ref}
\balance

\end{document}